\documentclass[10pt,twocolumn,letterpaper]{article}

\usepackage{iccv}
\usepackage{times}
\usepackage{epsfig}
\usepackage{graphicx}
\usepackage{amsmath}
\usepackage{amssymb}

\usepackage[pagebackref=true,breaklinks=true,letterpaper=true,colorlinks,bookmarks=false]{hyperref}

\iccvfinalcopy 

\def\iccvPaperID{1407} 

\ificcvfinal\fi
\begin{document}


\title{TURN TAP: Temporal Unit Regression Network for Temporal Action Proposals}

\author{Jiyang Gao$^{1*}$ \quad Zhenheng Yang$^{1*}$ \quad Chen Sun$^2$  \quad Kan Chen$^1$ \quad Ram Nevatia$^1$ \\
$^1$University of Southern California \qquad $^2$Google Research \\
{\tt\small \{jiyangga, zhenheny, kanchen, nevatia\}@usc.edu,\quad chensun@google.com} 
}

\maketitle

\newcommand\blfootnote[1]{%
  \begingroup
  \renewcommand\thefootnote{}\footnote{#1}%
  \addtocounter{footnote}{-1}%
  \endgroup
}

\blfootnote{$*$ indicates equal contributions.}

\begin{abstract}
  Temporal Action Proposal (TAP) generation is an important problem, as fast and accurate extraction of semantically important (e.g. human actions) segments from untrimmed videos is an important step for large-scale video analysis. We propose a novel Temporal Unit Regression Network (TURN) model. There are two salient aspects of TURN: (1) TURN jointly predicts action proposals and refines the temporal boundaries by temporal coordinate regression; (2) Fast computation is enabled by unit feature reuse: a long untrimmed video is decomposed into video units, which are reused as basic building blocks of temporal proposals. TURN outperforms the previous state-of-the-art methods under average recall (AR) by a large margin on THUMOS-14 and ActivityNet datasets, and runs at over 880 frames per second (FPS) on a TITAN X GPU. We further apply TURN as a proposal generation stage for existing temporal action localization pipelines, it outperforms state-of-the-art performance on THUMOS-14 and ActivityNet. 
\end{abstract}

\section{Introduction}


We address the problem of generating Temporal Action Proposals (TAP) in long untrimmed videos, akin to generation of object proposals in images for rapid object detection. As in the case for objects, the goal is to make action proposals have high precision and recall, while maintaining computational efficiency. 

\begin{figure}
\centering
\includegraphics[scale=0.54]{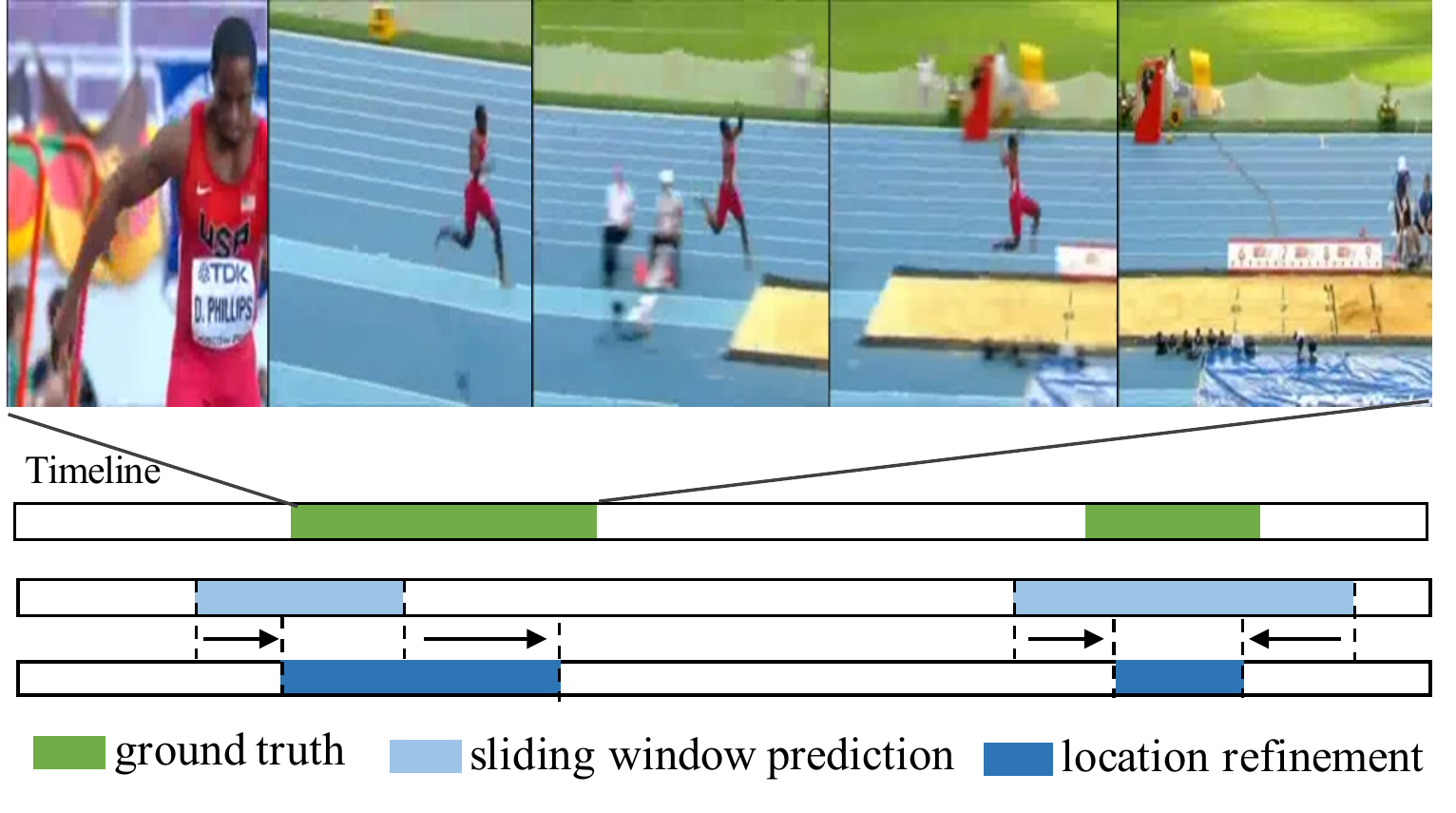}
\caption{ Temporal action proposal generation from a long untrimmed video. We propose a Temporal Unit Regression Network (TURN) to jointly predict action proposals and refine the location by temporal coordinate regression.}
\end{figure}




There has been considerable work in action classification task where a ``trimmed" video is classified into one of specified categories \cite{simonyan2014two, wang2015action}. There has also been work on localizing the actions in a longer, ``untrimmed" video \cite{gan2015devnet, Shou_2016_CVPR, Yuan_2016_CVPR, Yeung_2016_CVPR}, \emph{i.e.} temporal action localization. A straightforward way to use action classification techniques for localization is to use temporal sliding windows, however there is a trade-off between density of the sliding windows and computation time. Taking cues from the success of proposal frameworks in object detection tasks \cite{girshick2015fast, ren2015faster}, there has been recent work for generating temporal action proposals in videos \cite{Shou_2016_CVPR, escorcia2016daps, Heilbron_2016_CVPR} to improve the precision and accelerate the speed of temporal localization.

State-of-the-art methods \cite{Shou_2016_CVPR, Heilbron_2016_CVPR} formulate TAP generation as a binary classification problem (\emph{i.e.} action vs. background) and apply sliding window approach as well. Denser sliding windows usually would lead to higher recall rates at the cost of computation time. Instead of basing on sliding windows, Deep Action Proposals (DAPs) \cite{escorcia2016daps} uses a Long Short-term Memory (LSTM) network to encode video streams and infer multiple action proposals inside the streams. However, the performance of average recall (AR), which is computed by the average of recall at temporal intersection over union (tIoU) between 0.5 and 1, suffers at small number of predicted proposals compared with the sliding window based method \cite{Shou_2016_CVPR} \footnotemark.

To achieve high temporal localization accuracy and efficient computation cost, we propose to use temporal boundary regression. Boundary regression has been a successful practice for object localization, as in \cite{ren2015faster}. However, temporal boundary regression for actions has not been attempted in the past work. 

We present a novel method for fast TAP generation: Temporal Unit Regression Network (TURN). A long untrimmed video is first decomposed into short (\emph{e.g.} 16 or 32 frames) video \emph{units}, which serve as basic processing blocks. For each unit, we extract unit-level visual features using off-the-shelf models (C3D and two-stream CNN model are evaluated) to represent video units. Features from a set of contiguous units, called a clip, are pooled to create clip features. Multiple temporal scales are used to create a clip pyramid. To provide temporal context, clip-level features from the internal and surrounding units are concatenated. Each clip is then treated as a proposal candidate and TURN outputs a confidence score,  indicating whether it is an action instance or not. In order to better estimate the action boundary, TURN outputs two regression offsets for the starting time and ending time of an action in the clip. Non-maximum suppression (NMS) is then applied to remove redundant proposals. The source code is available at \url{https://github.com/jiyanggao/TURN-TAP}.
\footnotetext{Newly released evaluation results from DAPs authors show that SCNN-prop \cite{Shou_2016_CVPR} outperforms DAPs.}
\label{sec: metric correlation}
DAPs \cite{escorcia2016daps} and Sparse-prop \cite{Heilbron_2016_CVPR} use Average Recall vs. Average Number of retrieved proposals (AR-AN) to evaluate the TAP performance. There are two issues with AR-AN metric: (1) the correlation between AR-AN of TAP and mean Average Precision (mAP) of action localization was not explored
; (2) the average number of retrieved proposals is related to average video length of the test dataset, which makes AR-AN less reliable when evaluating across different datasets. Spatio-temporal action detection \cite{yu2015fast, Wang_2016_CVPR} used Recall vs. Proposal Number (R-N), however this metric does not take video lengths into consideration. 

There are two criteria for a good metric: (1) it should be capable of evaluating the performance of different methods on the same dataset effectively; (2) it should be capable of evaluating the performance of the same method across different datasets (generalization capability). We should expect better TAP would lead to better localization performance, using the same localizer. We propose a new metric, Average Recall vs. Frequency of retrieved proposals (AR-F), for TAP evaluation. In Section \ref{chap:4.2}, we validate that the proposed method satisfies the two criteria by quantitative correlation analysis between TAP performance and action localization performance.

We test TURN on THUMOS-14 and ActivityNet for TAP generation. Experimental results show that TURN outperforms the previous state-of-the-art methods \cite{escorcia2016daps, Shou_2016_CVPR} by a large margin under AR-F and AR-AN. For run-time performance, TURN runs at over 880 frames per second (FPS) with C3D features and 260 FPS with flow CNN features on a single TITAN X GPU. We further plug TURN as a proposal generation step in existing temporal action localization pipelines, and observe an improvement of mAP from state-of-the-art 19\% to 25.6\% (at tIoU=0.5) on THUMOS-14 by changing only the proposals. State-of-the-art localization performance is also achieved on ActivityNet. We show state-of-the-art performance on generalization capability by training TURN on THUMOS-14 and transfer it to ActivityNet without fine-tuning, strong generalization capability is also shown by test TURN across different subsets in ActivityNet without fine-tuning. 

In summary, our contributions are four-fold:

(1) We propose a novel architecture for temporal action proposal generation using temporal coordinate regression.

(2) Our proposed method achieves high efficiency (\textgreater800 fps) and outperforms previous state-of-the-art methods by a large margin.

(3) We show state-of-the-art generalization performance of TURN across different action datasets without dataset specific fine-tuning.

(4) We propose a new metric, AR-F, to evaluate the performance of TAP and compare AR-F with AR-AN and AR-N by quantitative analysis.

\section{Related Work}

\textbf{Temporal Action Proposal.} Sparse-prop \cite{Heilbron_2016_CVPR} proposes the use of STIPs \cite{laptev2005space} and dictionary learning for class-independent proposal generation. S-CNN \cite{Shou_2016_CVPR} presents a two-stage action localization system, in which the first stage is temporal proposal generation, and shows the effectiveness of temporal proposals for action localization. S-CNN's proposal network is based on fine-tuning 3D convolutional networks (C3D) \cite{tran2015learning} to binary classification task. DAPs \cite{escorcia2016daps} adopts LSTM networks to encode a video stream and produce proposals inside the video stream. 

\textbf{Temporal Action Localization.} Based on the progress of action classification, temporal action localization has been received much attentions recently. Ma \emph{et al.} \cite{Ma_2016_CVPR}  address the problem of early action detection. They propose to train a LSTM network with ranking loss and merge the detection spans based on the frame-wise prediction scores generated by the LSTM. Singh \emph{et al.} \cite{Singh_2016_CVPR} extend two-stream \cite{simonyan2014two} framework to multi-stream bi-directional LSTM networks and achieved state-of-the-art performance on MPII-Cooking dataset \cite{rohrbach2012database}. Sun \emph{et al.} \cite{sun2015temporal} transfer knowledge from web images to address temporal localization in untrimmed web videos. S-CNN \cite{Shou_2016_CVPR} presents a two-stage action localization framework: first using proposal networks to generate temporal proposals and then score the proposals with localization networks, which is trained with classification and localization loss.

\begin{figure*}
\centering
\includegraphics[scale=0.6]{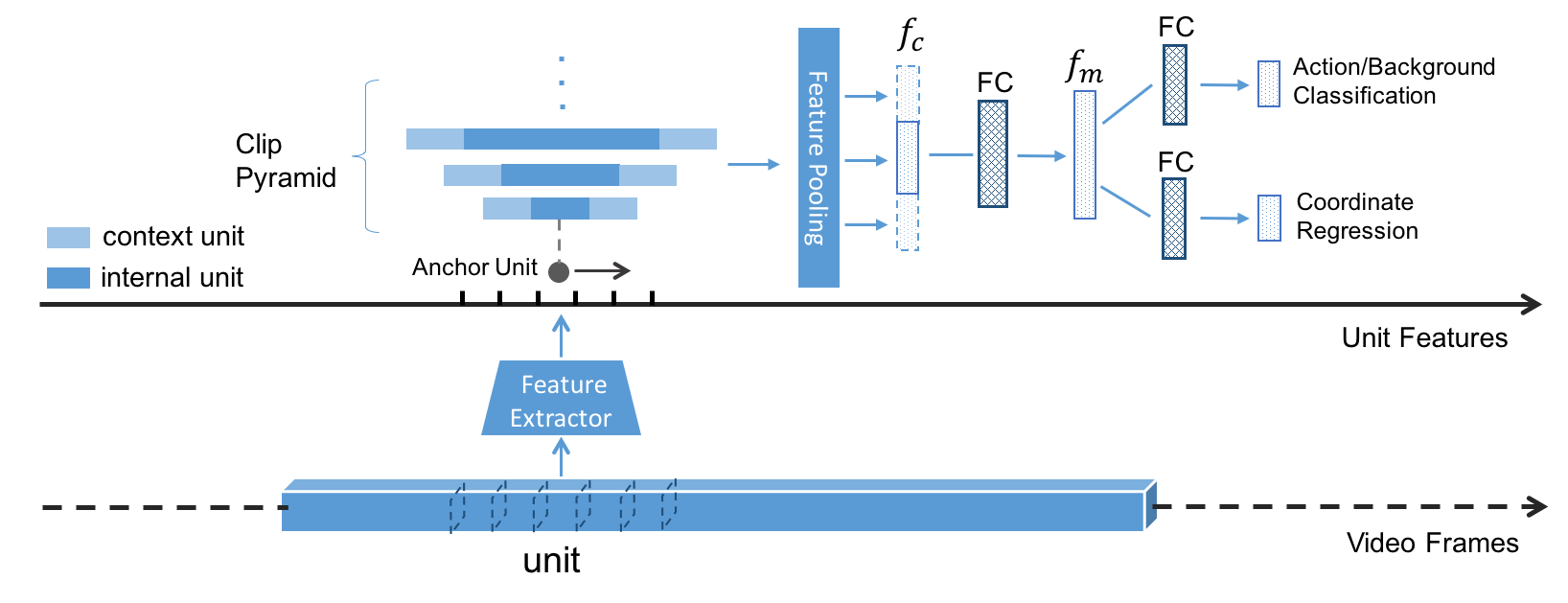}
\caption{ Architecture of Temporal Unit Regression Network (TURN). A long video is decomposed into short \emph{video units}, and CNN features are calculated for each unit. Features from a set of contiguous units, called a \emph{clip}, are pooled to create clip features. Multiple temporal scales are used to create a clip pyramid at an anchor unit. TURN takes a clip as input, and outputs a confidence score, indicating whether it is an action instance or not, and two regression offsets of start and end times to refine the temporal action boundaries.}
\end{figure*}

\textbf{Spatio-temporal Action Localization.}
A handful of efforts have been seen in spatio-temporal action localization. Gkioxari \emph{et al.} \cite{gkioxari2015finding} extract proposals from RGB images with SelectiveSearch \cite{uijlings2013selective} and then apply R-CNN \cite{Girshick_2014_CVPR} on both RGB and optical flow images for action detection. Weinzaepfel \emph{et al.} \cite{weinzaepfel2015learning} replace SelectiveSearch \cite{uijlings2013selective} with EdgeBoxes
\cite{zitnick2014edge}. Mettes \emph{et al.} \cite{mettes2016spot} propose to use sparse points as supervision for action detection to save tedious annotation work.

\textbf{Object Proposals and Detection.} 
Object proposal generation methods can be classified into two types based the features they use. One relies on hand-crafted low-level visual features, such as SelectiveSearch \cite{uijlings2013selective} and Edgebox \cite{zitnick2014edge}. R-CNN \cite{Girshick_2014_CVPR} and Fast R-CNN \cite{girshick2015fast} are built on this type of proposals. The other type is based on deep ConvNet feature maps, such as RPNs \cite{ren2015faster}, which introduces the use of anchor boxes and spatial regression for object proposal generation. YOLO \cite{Redmon_2016_CVPR} and SSD \cite{liu2015ssd} divide images into grids and regress object bounding boxes based on the grid cells. Bounding box coordinate regression is a common design shared in second type of object proposal frameworks. Inspired by object proposals, we adopt temporal regression in action proposal generation task.


\section{Methods}

In this section, we will describe the Temporal Unit Regression Network (TURN) and the training procedure.  

\subsection{Video Unit Processing} \label{sec:unit processing}
As we discussed before, the large-scale nature of video proposal generation requires the solution to be computationally efficient. Thus, extracting visual feature for the same window or overlapped windows repeatedly should be avoided. To accomplish this, we use \emph{video units} as the basic processing units in our framework. A video $V$ contains $T$ frames, $V=\{t_i\}_1^T$, and is divided into $T/n_u$ consecutive video units , where $n_u$ is the frame number of a unit. A unit is represented as $u=\{t_i\}_{s_f}^{s_f+n_u}$, where $s_f$ is the starting frame, $s_f+n_u$ is the ending frame. Units are not overlapped with each other.  

Each unit is processed by a visual encoder $E_v$ to get a unit-level representation $f_u=E_v(u)$. In our experiments, C3D \cite{tran2015learning}, optical flow based CNN model and RGB image CNN model \cite{simonyan2014two} are investigated. Details are given in Section \ref{chap:4.2}.

\subsection{Clip Pyramid Modeling}
A \emph{clip} (\emph{i.e. window}) $c$ is composed of units, $c=\{u_j\}_{s_u}^{s_u+n_c}$, where $s_u$ is the index of starting unit and $n_c$ is the number of units inside $c$. $e_u=s_u+n_c$ is the index of ending unit, and $\{u_j\}_{s_u}^{e_u}$ is called \emph{internal units} of $c$. Besides the internal units, \emph{context units} for $c$ are also modeled. $\{u_j\}_{s_u-n_{ctx}}^{s_u}$ and  $\{u_j\}_{e_u}^{e_u+n_{ctx}}$ are the context before and after $c$ respectively, $n_{ctx}$ is the number of units we consider for context. Internal feature and context feature are pooled from unit features separately by a function $P$. The final feature $f_c$ for a clip is the concatenation of context features and the internal features; $f_c$ is given by
\begin{equation*}
f_c=P(\{u_j\}_{s_u-n_{ctx}}^{s_u}) \parallel ~P(\{u_j\}_{s_u}^{e_u})\parallel ~P(\{u_j\}_{e_u}^{e_u+n_{ctx}})
\end{equation*}
where $\parallel$ represents vector concatenation and mean pooling is used for $P$. We scan an untrimmed video by building window pyramids at each unit position, \emph{i.e.} an \emph{anchor unit}. A clip pyramid $p$ consists of temporal windows with different temporal resolution, $p=\{c^{n_c}\}, n_c\in\{n_{c,1},n_{c,2},...\}$. Note that, although multi-resolution clips would have temporal overlaps, the clip-level features are computed from unit-level features, which are only calculated once. 

\subsection{Unit-level Temporal Coordinate Regression}
The intuition behind temporal coordinate regression is that human can infer the approximate start and end time of an action instance (\emph{e.g.} shooting basketball, swing golf) without watching the entire instance, similarly, neural networks might also be able to infer the temporal boundaries. Specifically, we design a unit regression model that takes a clip-level representation $f_c$ as input, and have two sibling output layers. The first one outputs a confidence score indicating whether the input clip is an action instance. The second one outputs temporal coordinate regression offsets. The regression offsets are
\begin{equation}
o_s=s_{clip}-s_{gt}, ~~ o_e=e_{clip}-e_{gt}
\end{equation}
where $s_{clip}$, $e_{clip}$ is the index of starting unit and ending unit of the input clip; $s_{gt}$, $e_{gt}$ is the index of starting unit and ending unit of the matched ground truth.

There are two salient aspects in our coordinate regression model. First, instead of regressing the temporal coordinates at frame-level, we adopt unit-level coordinate regression. As the basic unit-level features are extracted to encode $n_u$ frames, the feature may not be discriminative enough to regress the coordinates at frame-level. Comparing with frame-level regression, unit-level coordinate regression is easier to learn and more effective. Second, in contrast to spatial bounding box regression,  we don't use coordinate parametrization. We directly regress the offsets of the starting unit coordinates and the ending unit coordinates. The reason is that objects can be re-scaled in images due to camera projection, so the bounding box coordinates should be first normalized to some standard scale. However, actions' time spans can not be easily rescaled in videos.


\subsection{Loss Function}
For training TURN, we assign a binary class label (of being an action or not) to each clip (generated at each anchor unit). A positive label is assigned to a clip if: (1) the window clip with the highest temporal Intersection over Union (tIoU) overlaps with a ground truth clip; or (2) the window clip has tIoU larger than 0.5 with any of the ground truth clips. Note that, a single ground truth clip may assign positive labels to multiple window clips. Negative labels are assigned to non-positive clips whose tIoU is equal to 0.0 (\emph{i.e.} no overlap) for all ground truth clips. We design a multi-task loss $L$ to jointly train classification and coordinates regression.
\begin{equation}
L=L_{cls}+\lambda L_{reg}
\end{equation}
where $L_{cls}$ is the loss for action/background classification, which is a standard Softmax loss. $L_{reg}$ is for temporal coordinate regression and $\lambda$ is a hyper-parameter. The regression loss is 
\begin{equation}
L_{reg}=\frac{1}{N_{pos}}\sum_{i=1}^{N}l_i^*|(o_{s,i}-o^*_{s,i}) +(o_{e,i}-o^*_{e,i})|
\end{equation}
$L1$ distance is adopted. $l^*_i$ is the label, $1$ for positive samples and $0$ for background samples. $N_{pos}$ is the number of positive samples. The regression loss is calculated only for positive samples. 

During training, the background to positive samples ratio is set to be 10 in a mini-batch. The learning rate and batch size are set as 0.005 and 128 respectively. We use the Adam \cite{kingma2014adam} optimizer to train TURN.

\section{Evaluation}
In this section, we introduce the evaluation metrics, experimental setup and discuss the experimental results.

\subsection{Metrics}

  We consider three different metrics to assess the quality of TAP, the major difference is in the way to consider the retrieve number of proposals: Average Recall vs. Number of retrieved proposals (AR-N) \cite{yu2015fast,hosang2016makes}, Average Recall vs. Average Number of retrieved proposals (AR-AN) \cite{escorcia2016daps}, Average Recall vs. Frequency of retreived proposals (AR-F). 
 \textit{Average Recall} (AR) is calculated as a mean value of recall rate at tIoU between 0.5 and 1.

 \textbf{AR-N curve.} In this metric, the numbers of retrieved proposals (N) for all test videos are the same. This curve plots AR versus number of retrieved proposals. 
 
 \textbf{AR-AN curve.} In this metric, AR is calculated as a function of \textit{average number of retrieved proposals} (AN).  AN is calculated as:
    $\overline{\Theta} = \rho\overline{\Phi},  \rho\in(0,1]$. In which, $\overline{\Phi} = \frac{1}{n}\sum_{i=1}^{n} \Phi_i$ is the average number of all proposals of test videos. $\rho$ is the ratio of picked proposals to evaluate. $n$ is the number of test videos and $\Phi_i$ is the number of all proposals for each video. By scanning the ratio $\rho$ from 0 to 1, the number of retrieved proposals in each video varies from 0 to number of all proposals and thus the average number of retrieved proposals also varies.
 
 \textbf{AR-F curve.} This is the new metric that we propose. We measure average recall as a function of proposal frequency (F), which denotes the number of retrieved proposals per second for a video. For a video of length $l_i$ and proposal frequency of $F$, the retrieved proposal number of this video is $R_i = Fl_i$.
 
 We also report Recall@X-tIoU curve: recall rate at X with regard to different \textit{tIoU}. X could be number of retrieved proposals (N), average number of retrieved proposals (AN) and proposal frequency (F).

 For the evaluation of temporal action localization, we follow the traditional mean Average Precision (mAP) metric used in THUMOS-14 and ActivityNet. A prediction is regarded as positive only when it has correct category prediction and tIoU with ground truth higher than a threshold. We use the official evaluation toolkit provided by THUMOS-14 and ActivityNet.
 
\subsection{Experiments on THUMOS-14} 
\label{chap:4.2}
\textbf{Datasets.} The temporal action localization part of THUMOS-14 contains over 20 hours of videos from 20 sports classes. This part consists of 200 videos in validation set and 213 videos in test set. TURN model is trained on the validation set, as the training set of THUMOS-14 contains only trimmed videos. 

\textbf{Experimental setup.} \label{sec: setup} We perform the following experiments: (1) different temporal proposal evaluation metrics are compared; (2) the performance of TURN and other TAP generation methods are compared under evaluation metrics  (\emph{i.e} AR-F and AR-AN) mentioned above; (3) different TAP generation methods are compared on the temporal action localization task with the same localizer/classifier. Specifically, we feed the proposals into a localizer/classifier, which outputs the confidence scores of 21 classes (20 classes of action plus background). Two localizer/classifiers are adopted: (a) SVM classifiers: \textit{one-vs-all} linear SVM classifiers are trained for all 21 classes using C3D \textit{fc}6 features; (b) S-CNN localizer: the pre-trained localization network of S-CNN \cite{Shou_2016_CVPR} is adopted. \label{sec: exp setup}

For TURN model, the context unit number $n_{ctx}$ is 4, $\lambda$ is 2.0, the dimension of middle layer $f_m$ is 1000, temporal window pyramids is built with $\{1,2,4,8,16,32\}$ units. We test TURN with different unit sizes $n_u \in \{16, 32\}$, and different unit features, including C3D \cite{tran2015learning}, optical flow based CNN feature and RGB CNN feature \cite{simonyan2014two}. The NMS threshold is set to be 0.1 smaller than tIoU in evaluation. We implement TURN model in Tensorflow \cite{abadi2015tensorflow}.

\textbf{Comparison of different evaluation metrics.} 
To validate the effectiveness of different evaluation metrics, we
compare AR-F, AR-N, AR-AN by a correlation analysis
with localization performance (mAP). We generate
seven different sets of proposals, including random
proposals, slidinig windows and variants of S-CNN \cite{Shou_2016_CVPR} proposals (details are given in the supplementary material).
We then test the localization performance using
the proposals, as shown in Figure 3 (a)-(c). SVM classifiers
are used for localization.



\begin{figure}[h]
  \centering
    \includegraphics[width=0.48\textwidth]{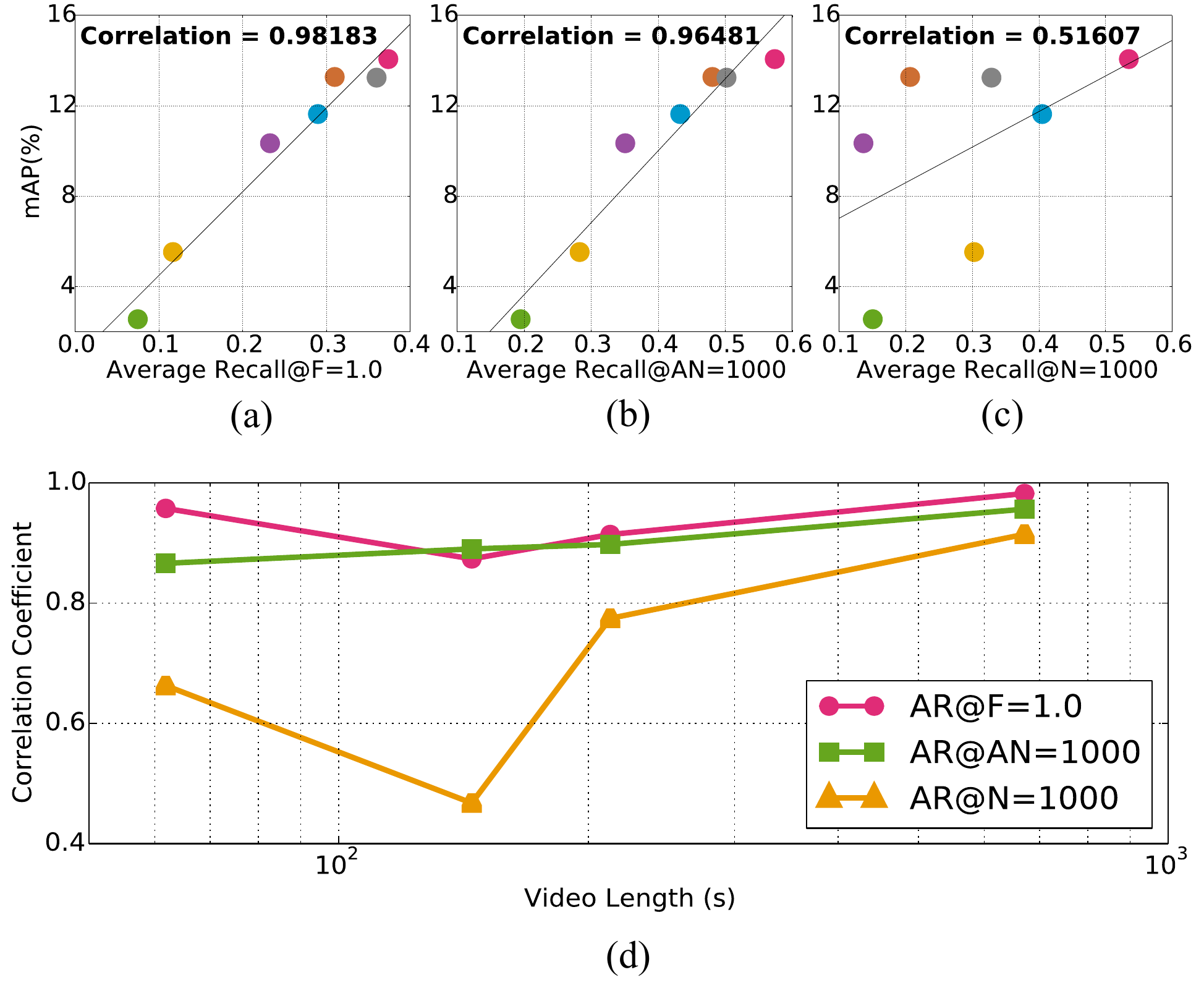}
    \caption{(a)-(c) show the correlation between temporal action localization performance and TAP performance  under different metrics. (d) shows correlation coefficient between temporal action localization and TAP performance versus video length on THUMOS-14 dataset.}
      \label{fig:correlation}
\end{figure}

A detailed analysis of correlation and video length is given in Figure \ref{fig:correlation} (d). The test videos are sorted by video lengths and then divided evenly into four groups. The average video length of the group is the x-axis, and y-axis represents the correlation coefficient between action localization performance and TAP performance of the group. Each point in \ref{fig:correlation} (d) represents the correlation of TAP and localization performance of one group under different evaluation metrics. As can be observed in Figure \ref{fig:correlation}, the correlation coefficient between mAP and AR-F is consistently higher than 0.9 at all video lengths. In contrast, correlation of AR-N and mAP is affected by video length distribution. Note that, AR-AN also shows a stable correlation with mAP, this is partially because the TAP generation methods we use generate proportional numbers of proposals to video length. 
 
To assess generalization, assume that we have two different datasets, $S_0$ and $S_1$, whose average number of all proposals are $\overline{\Phi}_0$ and $\overline{\Phi}_1$ respectively. As introduced before, average number of retrieved proposals $\overline{\Theta} = \rho\overline{\Phi},  \rho\in(0,1]$ is dependent on $\Phi$. When we compare AR at some certain $AN=\overline{\Theta}_x$ between $S_0$ and $S_1$, as $\overline{\Phi}_0$ and $\overline{\Phi}_1$ are different, we need to set different $\rho_0$ and $\rho_1$. It means that the ratios between retrieved proposals and all generated proposals are different for $S_0$ and $S_1$, which make the AR calculated for $S_0$ and $S_1$ at the same $AN=\overline{\Theta}_x$ can not be compared directly. For AR-F, the number of proposals retrieved is based on ``frequency", which is independent with the average number of all generated proposals.

In summary, AR-N cannot evaluate TAP performance effectively on the same dataset, as number of retrieved proposals should vary with video lengths. AR-AN cannot be used to compare TAP performance among different datasets, as the retrieval ratio depends on dataset's video length distribution, which makes the comparison unreasonable. AR-F satisfies both requirements. 


\textbf{Comparison of visual features.} We test TURN with three unit-level features to assess the effect of visual features on AR performance: C3D \cite{tran2015learning} features, RGB CNN features with temporal mean pooling and dense flow CNN \cite{xiong2016cuhk} features. The C3D model is pre-trained on Sports1m \cite{Karpathy_2014_CVPR}, all 16 frames in a unit are input into C3D and the output of $fc$6 layer is used as unit-level feature. For RGB CNN features, we  uniformly sample 8 frames from a unit, extract ``Flatten\_673" features using a ResNet \cite{He_2016_CVPR} model (pre-trained on training set of ActivityNet v1.3 dataset \cite{xiong2016cuhk}) and compute the mean of these 8 features as the unit-level feature. For dense flow CNN features, we sample $6$ consecutive frames at the center of a unit and calculate optical flow \cite{farneback2003two} between them. The flows are then fed into a BN-Inception model \cite{xiong2016cuhk,ioffe2015batch} that is pre-trained on training set of ActivityNet v1.3 dataset \cite{xiong2016cuhk}. The output of ``global pool" layer of BN-Inception is used as the unit-level feature.

\begin{figure}[]

  \centering
    \includegraphics[width=0.48\textwidth]{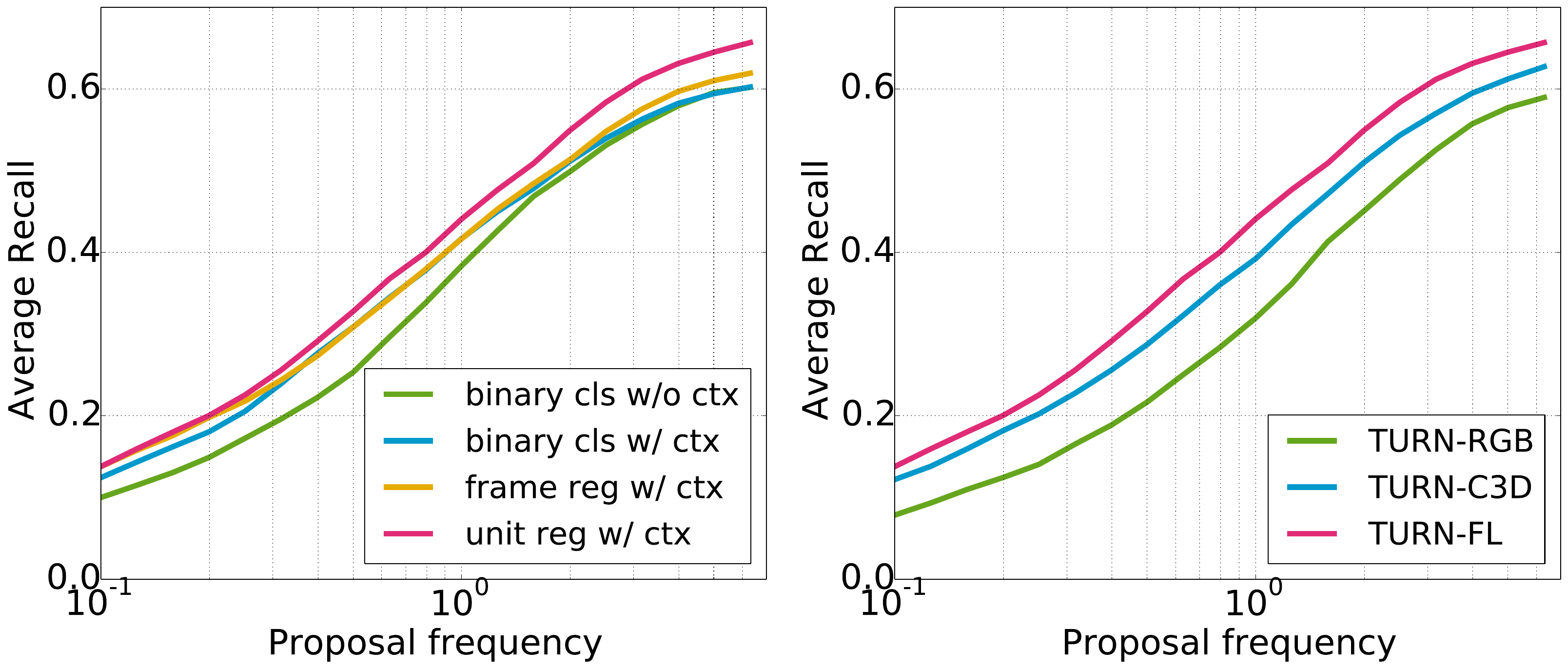}
    \caption{Comparison of TURN variants on THUMOS-14 dataset}
      \label{fig:compare_variant}
\end{figure}

As shown in Figure \ref{fig:compare_variant}, dense flow CNN feature (TURN-FL) gives the best results, indicating optical flow can capture temporal action information effectively. In contrast, RGB CNN features (TURN-RGB) show inferior performance and C3D (TURN-C3D) gives competitive performance.

\textbf{Temporal context and unit-level coordinate regression.} We compare four variants of TURN to show the effectiveness of temporal context and unit regression: (1) \textit{binary cls w/o ctx}: binary classification (no regression) without the use of temporal context, (2) \textit{binary cls w/ ctx}: binary classification (no regression) with the use of context, (3) \textit{frame reg w/ ctx}: frame-level coordinate regression with the use of context and (4) \textit{unit reg w/ ctx}: unit-level coordinate regression with the use of context (\emph{i.e.} our full model). The four variants are compared with AR-F curves. As shown in Figure \ref{fig:compare_variant}, temporal context helps to classify action and background by providing additional information. As shown in AR-F curve, \textit{unit reg w/ ctx} has higher AR than the other variants at all frequencies, indicating that unit-level regression can effectively refine the proposal location. Some TURN proposal results are shown in Figure \ref{fig:example}.

\begin{figure}[]
   \centering
    \includegraphics[width=0.48\textwidth]{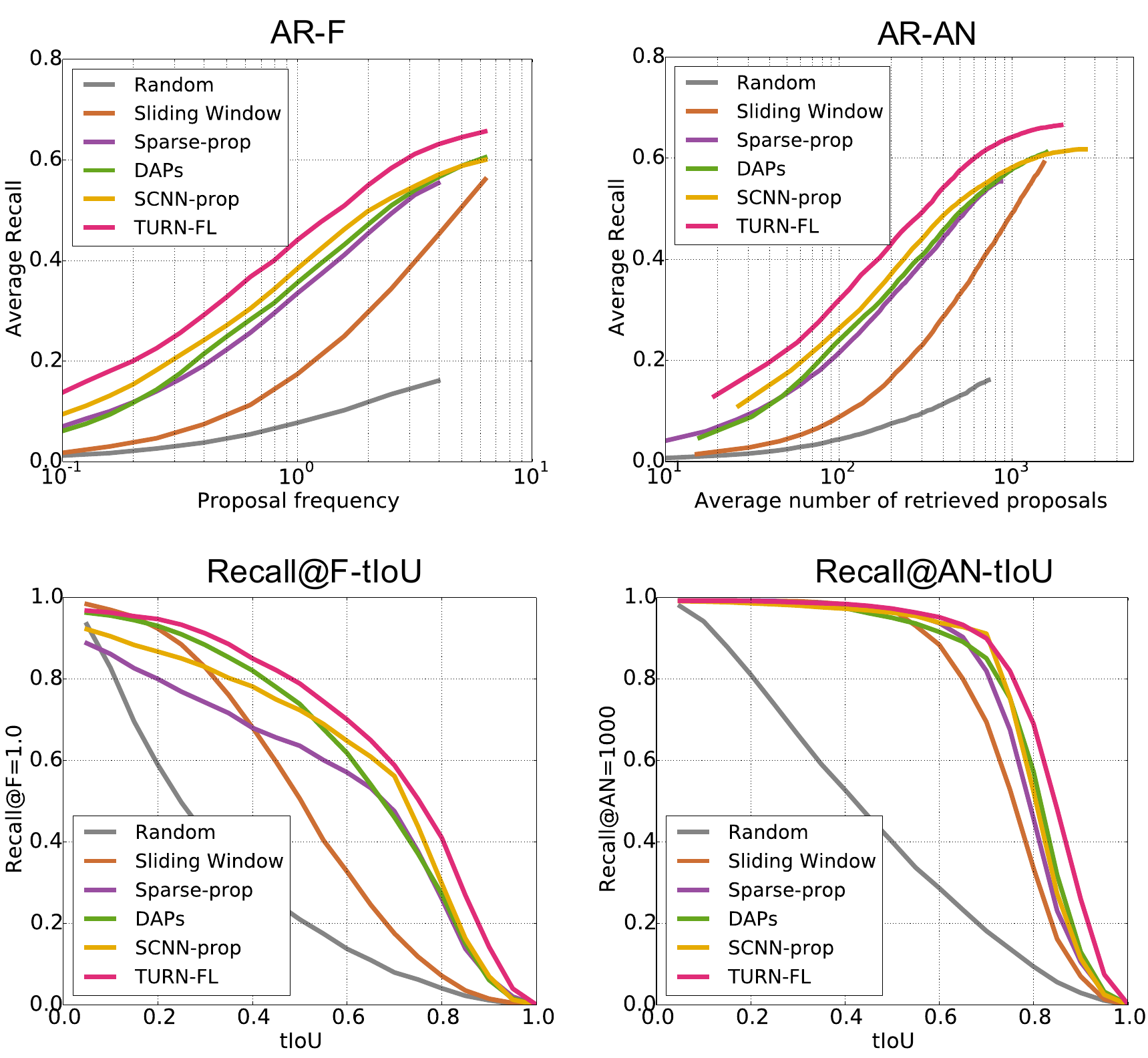}
    \caption{Proposal performance on THUMOS-14 dataset under 4 metrics: AR-F, AR-AN, Recall@F-tIoU, Recall@AN-tIoU. For AR-AN and Recall@AN-tIoU, we use the codes provided by \cite{escorcia2016daps}}
     \label{fig:compare_six}
\end{figure}

\begin{figure}[]

  \centering
    \includegraphics[width=0.5\textwidth]{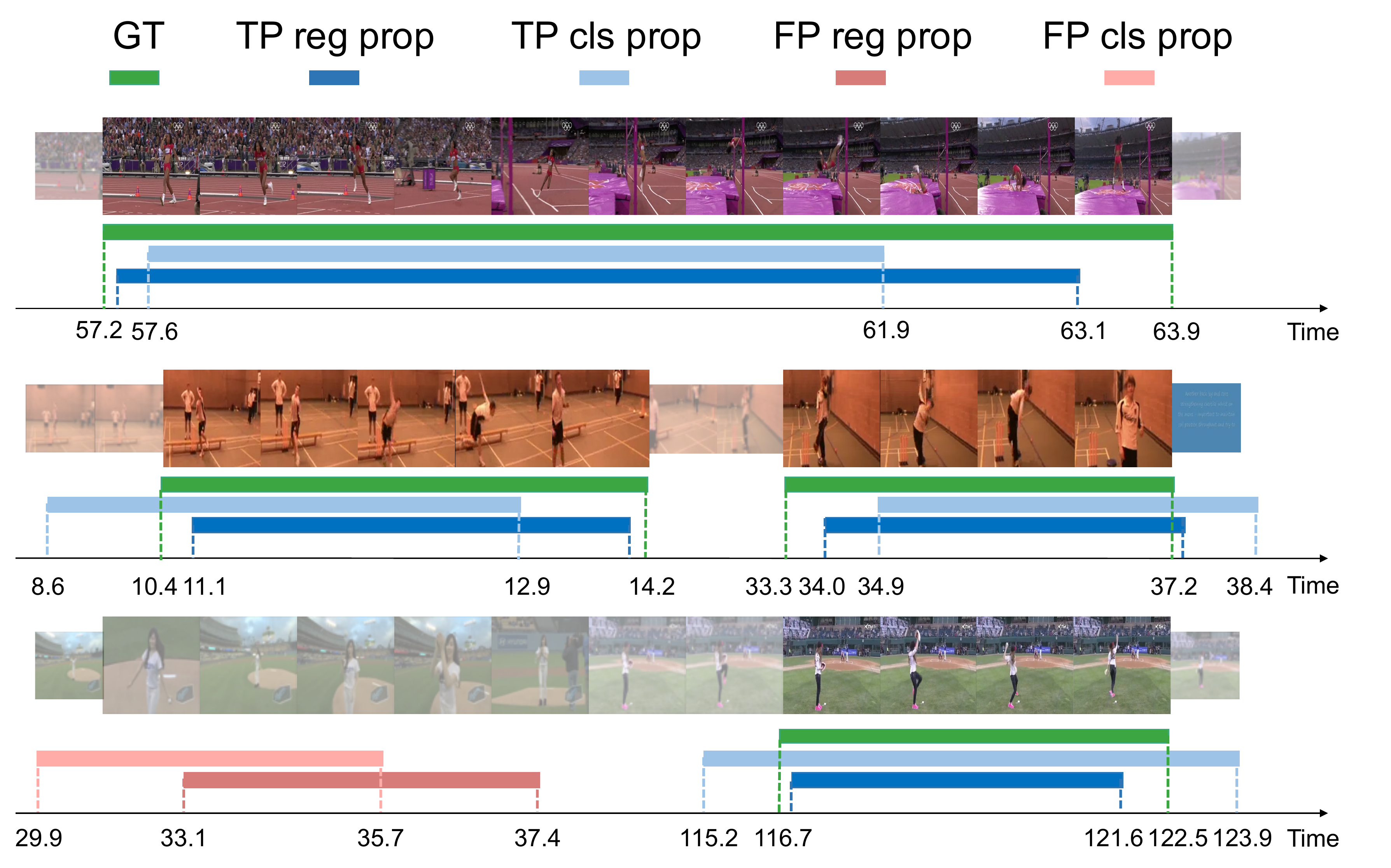}
    \caption{Qualitative examples of retrieved proposals by TURN on THUMOS-14 dataset. GT indicates ground truth. TP and FP indicate true positive and false positive respectively. ``reg prop" and ``cls prop" indicate regression proposal and classification proposal.}
      \label{fig:example}
\end{figure}

\textbf{Comparison with state-of-the-art.} We compare TURN with the state-of-the-art methods under AR-AN, AR-F, Recall@AN-tIoU, Recall@F-tIoU metrics. The TAP generation methods include DAPs \cite{escorcia2016daps}, SCNN-prop \cite{Shou_2016_CVPR}, Sparse-prop \cite{Heilbron_2016_CVPR}, \textit{sliding window}, and \textit{random proposals}. For DAPs, Sparse-prop and SCNN-prop, we plot the curves using the proposal results provided by the authors. ``Sliding window proposals" include all sliding windows of length from 16 to 512 overlapped by 75\%, each window is assigned with a random score. ``Random proposals" are generated by assigning random starting and ending temporal coordinates (ending temporal coordinate is larger than starting temporal coordinate), each random window is assigned with a random score. As shown in Figure \ref{fig:compare_six}, TURN outperforms the state-of-the-art consistently by a large margin under all four metrics.

\textbf{How unit size affects AR and run-time performance?}
The impact of unit size on AR and computation speed is evaluated with $n_u \in \{16,32\}$. We keep other hyper-parameters the same as in Section \ref{sec: setup}. Table \ref{tbl:fps} shows comparison of the three TURN variants (TURN-FL-16, TURN-FL-32, TURN-C3D-16) and three state-of-the-art TAP methods, in terms of recall (AR@F=1.0) and run-time (FPS) performance. We randomly select 100 videos from THUMOS-14 validation set and run TURN-FL-16, TURN-FL-32 and TURN-C3D-16 on a single Nvidia TITAN X GPU. The run-time of DAPs \cite{escorcia2016daps} and SCNN-prop \cite{Shou_2016_CVPR} are provided in \cite{escorcia2016daps}, which were tested on a TITAN X GPU and a GTX 980 GPU respectively. The hardware used in \cite{Heilbron_2016_CVPR} is not specified in the paper.
\begin{table}[h]
\centering
\caption{Run-time and AR Comparison on THUMOS-14.}
\label{tbl:fps}
\begin{tabular}{l|cc}
\hline
method      & AR@F=1.0 (\%) & FPS   \\ \hline
DAPs \cite{escorcia2016daps}        & 35.7     & 134.3 \\
SCNN-prop \cite{Shou_2016_CVPR}   & 38.3     & 60.0  \\
Sparse-prop \cite{Heilbron_2016_CVPR} & 33.3     & 10.2  \\ \hline 
TURN-FL-16     & \textbf{43.5}     & 129.4 \\
TURN-FL-32     & 42.4     & 260.6 \\
TURN-C3D-16     & 39.3     & \textbf{880.8} \\ \hline
\end{tabular}
\end{table}

As can be seen, there is a trade-off between AR and FPS: smaller unit size leads to higher recall rate, and also higher computational complexity. We consider unit size as temporal coordinate precision, for example,  unit size of 16 and 32 frames represent approximately half second and one second respectively. The major part of computation time comes from unit-level feature extraction. Smaller unit size leads to more number of units, which increases computation time; on the other hand, smaller unit size also increases temporal coordinate precision, which improves the precision of temporal regression. C3D feature is faster than flow CNN feature, but with a lower performance. Compared with state-of-the-art methods, we can see that TURN-C3D-16 outperforms current state-of-the-art AR performance, but accelerates computation speed by more than 6 times. TURN-FL-16 achieves the highest AR performance with competitive run-time performance.


\textbf{TURN for temporal action localization.} We feed proposal results of different TAP generation methods into the same temporal action localizers/classifiers to compare the quality of proposals. The value of mAP@tIoU=0.5 is reported in Table \ref{diff prop tbl}. TURN outperforms all other methods in both the SVM classifier and S-CNN localizer. Sparse-prop, SCNN-prop and DAPs all use C3D to extract features. It is worth noting that the localization results of four different proposals suit well with their proposal performance under AR-F metric in Figure \ref{fig:compare_six}: the methods that have better performance under AR-F achieve higher mAP in temporal action localization.

\begin{table}[h]\small
\centering
\caption{Temporal action localization performance (mAP \% @tIoU=0.5) evaluated on different proposals on THUMOS-14.}
\label{diff prop tbl}
\begin{tabular}{l|ccc}
\hline
\multicolumn{1}{l|}{} & \multicolumn{1}{l}{DAPs SVM\cite{escorcia2016daps}} & \multicolumn{1}{l}{Our SVM} & \multicolumn{1}{l}{S-CNN} \\ \hline
Sparse-prop\cite{Heilbron_2016_CVPR}          & 7.8                          & 8.1                         & 15.3                     \\ \hline
DAPs\cite{escorcia2016daps}                 & 13.9                         & 9.5                         & 16.3                     \\ \hline
SCNN-prop\cite{Shou_2016_CVPR}            & 7.6\protect\footnotemark                          & 14.0                        & 19.0                     \\ \hline
TURN-C3D-16                 & -                            & 16.4               & 22.5            \\ \hline
TURN-FL-16                 & -                            & \textbf{17.8}               & \textbf{25.6}            \\ \hline
\end{tabular}
\end{table}
\addtocounter{footnote}{0}
\footnotetext{ This number should be higher, as DAPs authors adopted an incorrect frame rate when using S-CNN proposals.}

A more detailed comparison of state-of-the-art localization methods is given in Table \ref{diff iou tbl}. It can be seen that, by applying TURN with linear SVM classifiers for action localization, we achieve comparable performance with the state-of-the-art methods. By further incorporating S-CNN localizer, we outperform all other methods by a large margin at all tIoU thresholds. The experimental results prove the high-quality of TURN proposals.

\begin{table}[h]\small
\centering
\caption{Temporal action localization performance (mAP \%) comparison at different tIoU thresholds on THUMOS-14.}
\label{diff iou tbl}
\begin{tabular}{l|ccccc}
\hline
tIoU     & 0.1           & 0.2           & 0.3           & 0.4           & 0.5           \\ \hline
Oneata \emph{et al.}\cite{oneata2014lear} & 36.6          & 33.6          & 27.0          & 20.8          & 14.4          \\ 
Yeung \emph{et al.}\cite{Yeung_2016_CVPR}  & 48.9          & 44.0          & 36.0          & 26.4          & 17.1          \\ 
Yuan \emph{et al.} \cite{Yuan_2016_CVPR}  & 51.4          & 42.6          & 33.6          & 26.1          & 18.8          \\ 
S-CNN \cite{Shou_2016_CVPR}         & 47.7          & 43.5          & 36.3          & 28.7          & 19.0          \\ \hline
TURN-C3D-16 + SVM            & 46.4      & 41.5          & 34.3          & 24.9          & 16.4       \\ 
TURN-FL-16 + SVM            & 48.3      & 43.2          & 35.1          & 26.2          & 17.8       \\ 
TURN-C3D-16 +S-CNN          & 48.8      & 45.5          & 40.3          & 31.5          & 22.5 \\ 
TURN-FL-16 + S-CNN               & \textbf{54.0} & \textbf{50.9} & \textbf{44.1} & \textbf{34.9} & \textbf{25.6}\\ \hline
\end{tabular}
\end{table}

TURN helps action localization on two aspects: (1) TURN serves as the first stage of a localization pipeline (\emph{e.g.} S-CNN, SVM) to generate high-quality TAP, and thus increases the localization performance; (2) TURN accelerates localization pipelines by filtering out many background segments, thus reducing the unnecessary computation.

\vspace{-1.5mm}
\subsection{Experiments on ActivityNet}
\textbf{Datasets.} ActivityNet datasets provide rich and diverse action categories. There are three releases of ActivityNet dataset: v1.1, v1.2 and v1.3.
All three versions define a 5-level hierarchy of action classes. Nodes on higher level represent more abstract action categories. For example, the node ``Housework" on level-3 has child nodes ``Interior cleaning", ``Sewing, repairing, \& maintaining textiles" and ``Laundry" on level-4. From the hierarchical action categories definition, a subset can be formed by including all action categories that belong to a certain node.

\textbf{Experiment setup.} 
To compare with previous work, we do experiments on v1.1 (on subsets of ``Works" and ``Sports") for temporal action localization \cite{caba2015activitynet, Yeung_2016_CVPR}, v1.2 for proposal generalization capability following  the same evaluation protocol as in \cite{escorcia2016daps}. On v1.3, we design a different experimental setup to test TURN's cross-domain generalization capability: four subsets having distinct semantic meanings are selected, including ``Participating in Sports, Exercise, or Recreation", ``Vehicles", ``Housework" and ``Arts and Entertainment". We also check that the action categories in different subsets are not semantically related: for example, "archery", "dodge ball" in ``Sports" subset, "changing car wheels", "fixing bicycles" in ``Vehicles" subset, "vacuuming floor", "cleaning shoes" in ``Housework" subset, "ballet", "playing saxophone" in ``Arts" subset. 


The evaluation metrics include AR@AN curve for temporal action proposal and mAP for action localization. AR@F=1.0 is reported for comparing proposal performance on different subsets. The validation set is used for testing as the test set is not publicly available.

To train TURN, we set the number of frames in a unit $n_u$ to be 16, the context unit number $n_{ctx}$ to be 4, $L$ to be 6 and $\lambda$ to be 2.0. We build the temporal window pyramid with $\{2,4,8,16,32,64,128\}$ number of units. The NMS threshold is set to be 0.1 smaller than tIoU in evaluation. For the temporal action localizer, SVM classifiers are trained with two-stream CNN features in ``Sports"  and ``Works" subsets.

\textbf{Generalization capability of TURN.}
One important property of TAP is the expectation to generalize beyond the categories it is trained on. 

\begin{figure}[]

  \centering
    \includegraphics[width=0.46\textwidth]{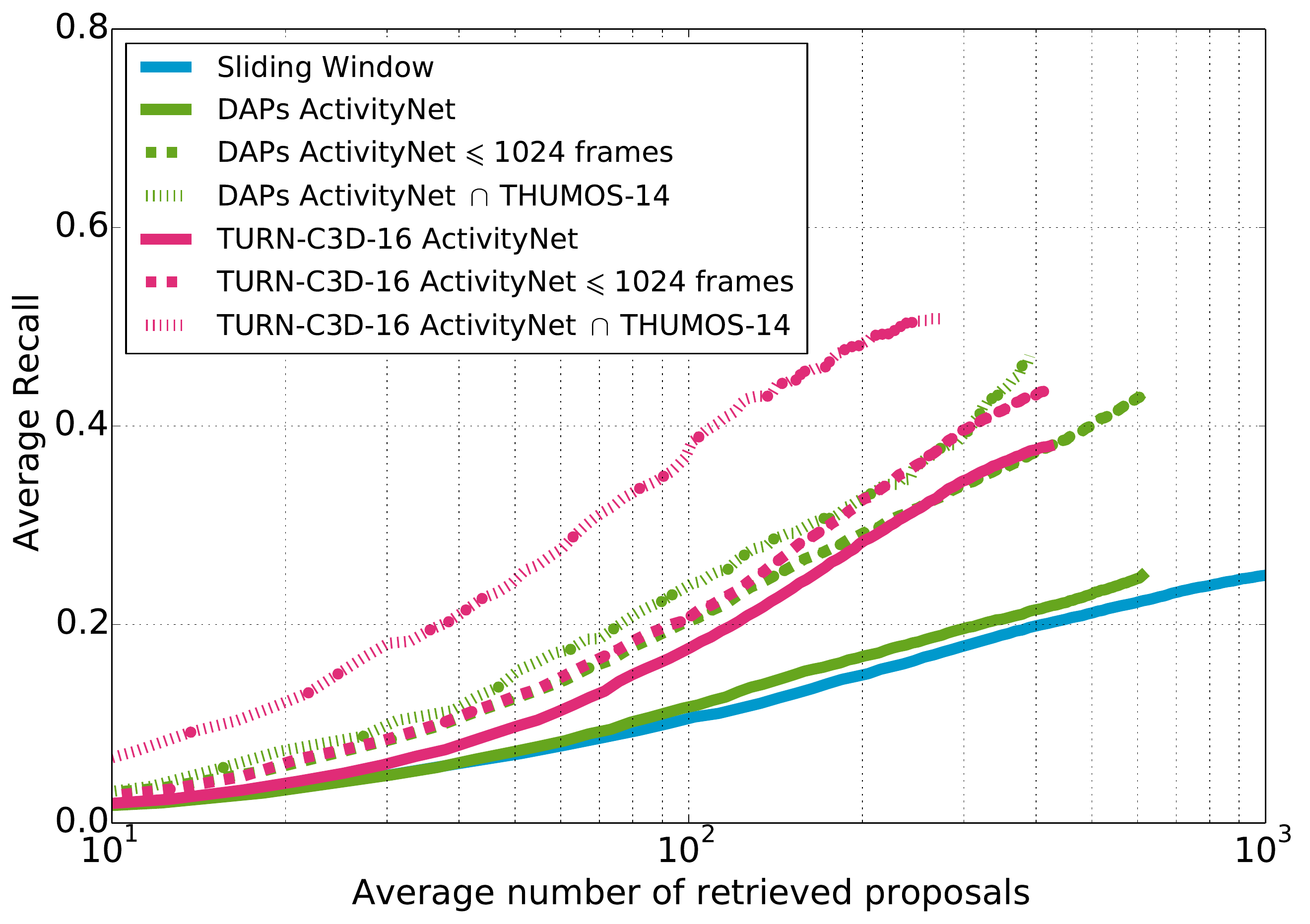}
    \caption{Comparison of generalizability on ActivityNet v1.2 dataset}
      \label{fig:generalization}
\end{figure}

On ActivityNet v1.2, we follow the same evaluation protocol from \cite{escorcia2016daps}: model trained on THUMOS-14 validation set and tested in three different sets of ActivityNet v1.2: the whole set of \textit{ActivityNet v1.2} (all 100 categories), \textit{ActivityNet v1.2} $\cap$ THUMOS-14 (on 9 categories shared between the two) and \textit{ActivityNet v1.2} $\leqslant$ 1024 frames (videos with unseen categories with annotations up to 1024 frames). To avoid any possible dataset overlap and enable direct comparison, we use C3D (pre-trained on Sports1M) as feature extractor, the same as DAPs did. As shown in Figure \ref{fig:generalization}, TURN has better generalization capability in all three sets.


\begin{table}[h]\footnotesize
\centering
\caption{Proposal generalization performance (AR@F=1.0 \%) of TURN-C3D-16 on different subsets of ActivityNet.}
\label{tbl:generalization}
\begin{tabular}{l|cccc}
\hline
\multicolumn{1}{l|}{} & \multicolumn{1}{l}{Arts} & \multicolumn{1}{l}{Housework} & \multicolumn{1}{l}{Vehicles} & Sports \\ \hline
Sliding Windows   &24.44    &27.63    &27.59    &25.72 \\ \hline
Arts (23; 685)                  & 44.30                    & 44.38                         & 40.85                         & 38.43  \\
Housework (10; 373)             & 40.27                    & 44.30                         & 38.65                        & 36.54  \\
Vehicles (5; 238)              & 38.43                    & 40.05                         & 42.22                        & 30.70  \\
Sports (26; 1294)                & 43.26                    & 43.58                         & 41.40                        & 46.62  \\ 
Ensemble (64; 2590)        &45.30      &48.12     &42.33      &46.72 \\ \hline 
\end{tabular}
\end{table}

On ActivityNet v1.3, we implement a different setup for evaluating generalization capability on subsets that contain semantically distinct actions: (1) we train TURN on one subset and test on the other three subsets, (2) we train on the ensemble of all 4 subsets and test on each subset. TURN is trained with C3D unit features, to avoid any overlap of training data. We also report performance of sliding windows (lengths of 32, 64, 128, 256, 512, 1024 and 2048, overlap 50\% ) in each subset. Average recall at frequency 1.0 (AR@F=1.0) are reported in Table \ref{tbl:generalization}. The left-most column lists subsets used for training. The numbers of action classes and training videos with each subset are shown in brackets. The top row lists subsets for test. The off-diagonal elements indicate that the training data and test data are from different subsets; the diagonal elements indicate the training data and test data are from the same subsets.

As can be seen in Table \ref{tbl:generalization}, the overall generalization capability is strong. Specifically, the generalization capability when training on ``Sports" subset is the best compared with other subsets, which may indicate that more training data would lead to better generalization performance. The ``Ensemble" row shows that using training data from other subsets would not harm the performance of each subset. 




\textbf{TURN for temporal action localization.}
Temporal action localization performance is evaluated and compared on ``Works" and ``Sports" subsets of ActivityNet v1.1. TURN trained with dense flow CNN features is used for comparison. On v1.1, TURN-FL-16 proposal is fed into \textit{one-vs-all} SVM classifiers which trained with two-stream CNN features. From the results shown in Table \ref{tbl: tal of anet}, we can see that TURN proposals improve localization performance.

\begin{table}[h]\small
\centering
\caption{Temporal action localization performance (mAP\% @tIoU=0.5) on ActivityNet v1.1}
\label{tbl: tal of anet}
\begin{tabular}{l|cccc}
\hline
Subsets & {\cite{caba2015activitynet}} & {\cite{Yeung_2016_CVPR}} & Sliding Windows & TURN-FL-16 \\ \hline
Sports  & 33.2   & 36.7   & 27.3    & \textbf{37.1}           \\ \hline
Work    & 31.1   & 39.9   & 29.6      &   \textbf{41.2}             \\ \hline
\end{tabular}
\end{table}

\section{Conclusion}

We presented a novel and effective Temporal Unit Regression Network (TURN) for fast TAP generation. We proposed a new metric for TAP: Average Recall-Proposal Frequency (AR-F). AR-F is robustly correlated with temporal action localization performance and it allows performance comparison among different datasets. TURN can runs at over 880 FPS with the state-of-the-art AR performance. 
TURN is robust on different visual features, including C3D and dense flow CNN features. 
We showed the effectiveness of TURN as a proposal generation stage in localization pipelines on THUMOS-14 and ActivityNet. 

{\small
\bibliographystyle{ieee}
\bibliography{egbib}
}

\end{document}